\documentclass[10pt,twocolumn,letterpaper]{article}

\usepackage{wacv}
\usepackage{times}
\usepackage{epsfig}
\usepackage{graphicx}
\usepackage{amsmath}
\usepackage{amssymb}
\usepackage{multirow}
\usepackage{hyperref}
\usepackage{subfig}
\usepackage{xcolor}
\usepackage{cite}
\usepackage{array}
\usepackage[noabbrev,capitalize]{cleveref}

\usepackage{authblk}

\wacvfinalcopy 

\def\wacvPaperID{453} 

\ifwacvfinal\fi
\begin{document}

\title{Component-based Attention for Large-scale Trademark Retrieval}


\author[1]{Osman Tursun \thanks{Corresponding author: w.tuerxun@qut.edu.au}}

\author[1]{Simon Denman}
\author[1]{Sabesan Sivapalan}
\author[1]{Sridha Sridharan}
\author[1]{Clinton Fookes}
\author[2]{Sandra Mau}
\affil[1]{Image and Video Research Laboratory, SAIVT, Queensland University of Technology}
\affil[2]{TrademarkVision}

\maketitle
\ifwacvfinal\thispagestyle{empty}\fi

\begin{abstract}
The demand for large-scale trademark retrieval (TR) systems has significantly increased to combat the rise in international trademark infringement. Unfortunately, the ranking accuracy of current approaches using either hand-crafted or pre-trained deep convolution neural network (DCNN) features is inadequate for large-scale deployments. We show in this paper that the ranking accuracy of TR systems can be significantly improved by incorporating hard and soft attention mechanisms, which direct attention to critical information such as figurative elements and reduce attention given to distracting and uninformative elements such as text and background. Our proposed approach achieves state-of-the-art results on a challenging large-scale trademark dataset.
\end{abstract}


\section{Introduction}
\label{sec:intr}
A trademark or logo is a representative figure of a company or an organization, which needs to be registered in patent offices to protect it from infringements and piracy. However, the trademark registration procedure is a lengthy and time-consuming process, especially because of the rapid increase in the total number of trademark applications and registrations. Efficient trademark registration requires an automated trademark retrieval (TR) system that returns, with high accuracy, all trademarks that are similar or related to a given trademark query.

\begin{figure}[!t]
	\centering
	\subfloat[Shape]{
		\includegraphics[width=0.065\textwidth]{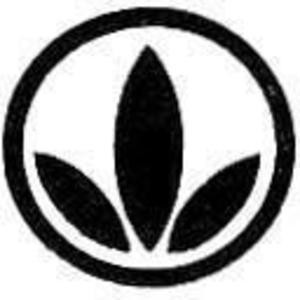}
		\includegraphics[width=0.065\textwidth]{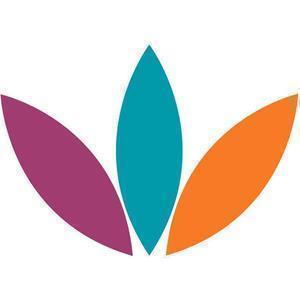}
	}
	\subfloat[Phonetic]{
		\includegraphics[width=0.07\textwidth]{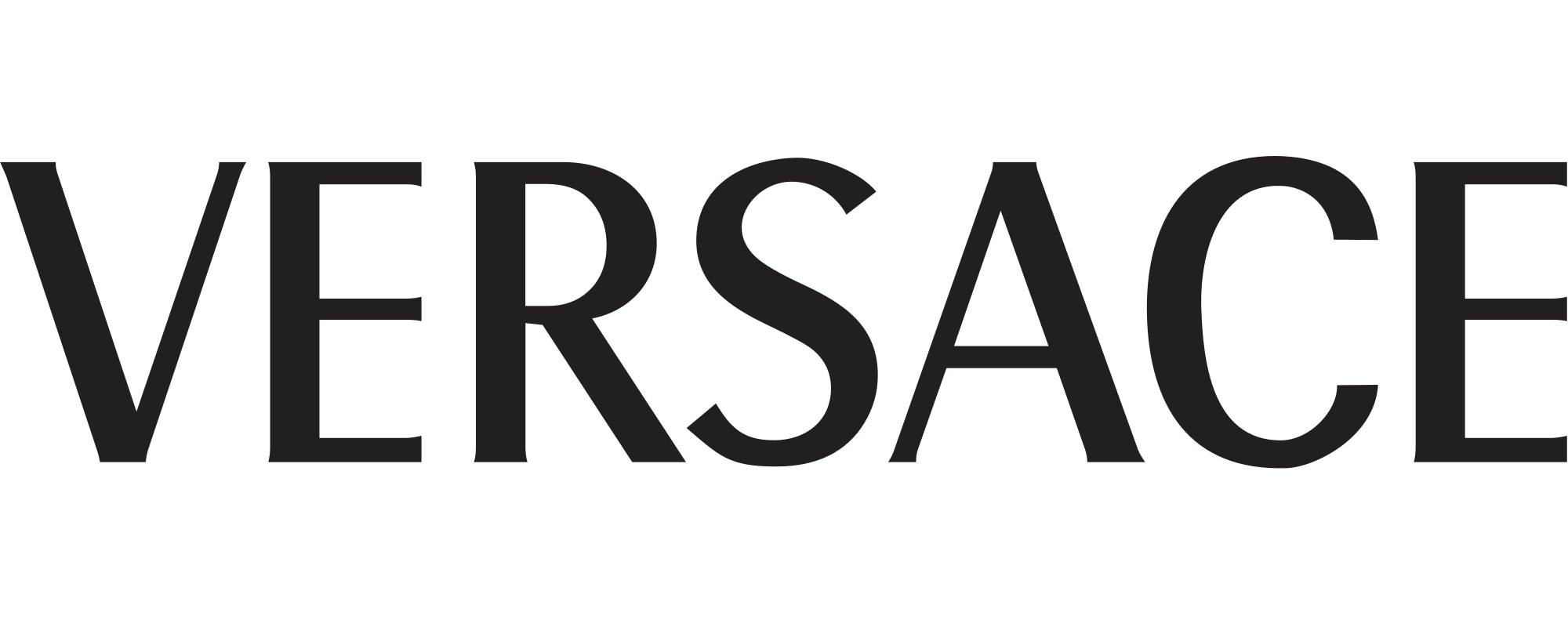}
		\includegraphics[width=0.07\textwidth]{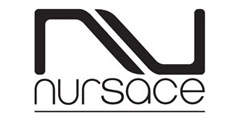}
	}
	\subfloat[Semantic]{
		\includegraphics[width=0.065\textwidth]{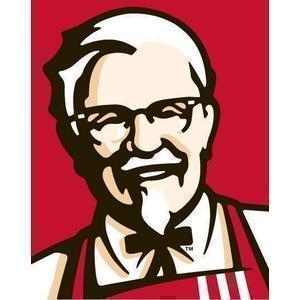}
		\includegraphics[width=0.065\textwidth]{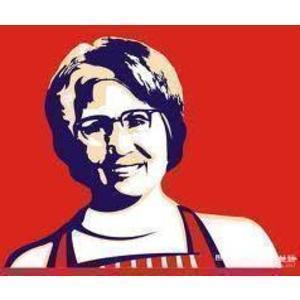}
	}
	\\
	\subfloat[Layout]{
		\includegraphics[width=0.07\textwidth]{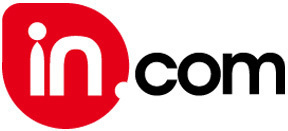}
		\includegraphics[width=0.07\textwidth]{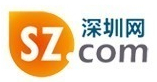}
	}
	\subfloat[Texture]{
		\includegraphics[width=0.065\textwidth]{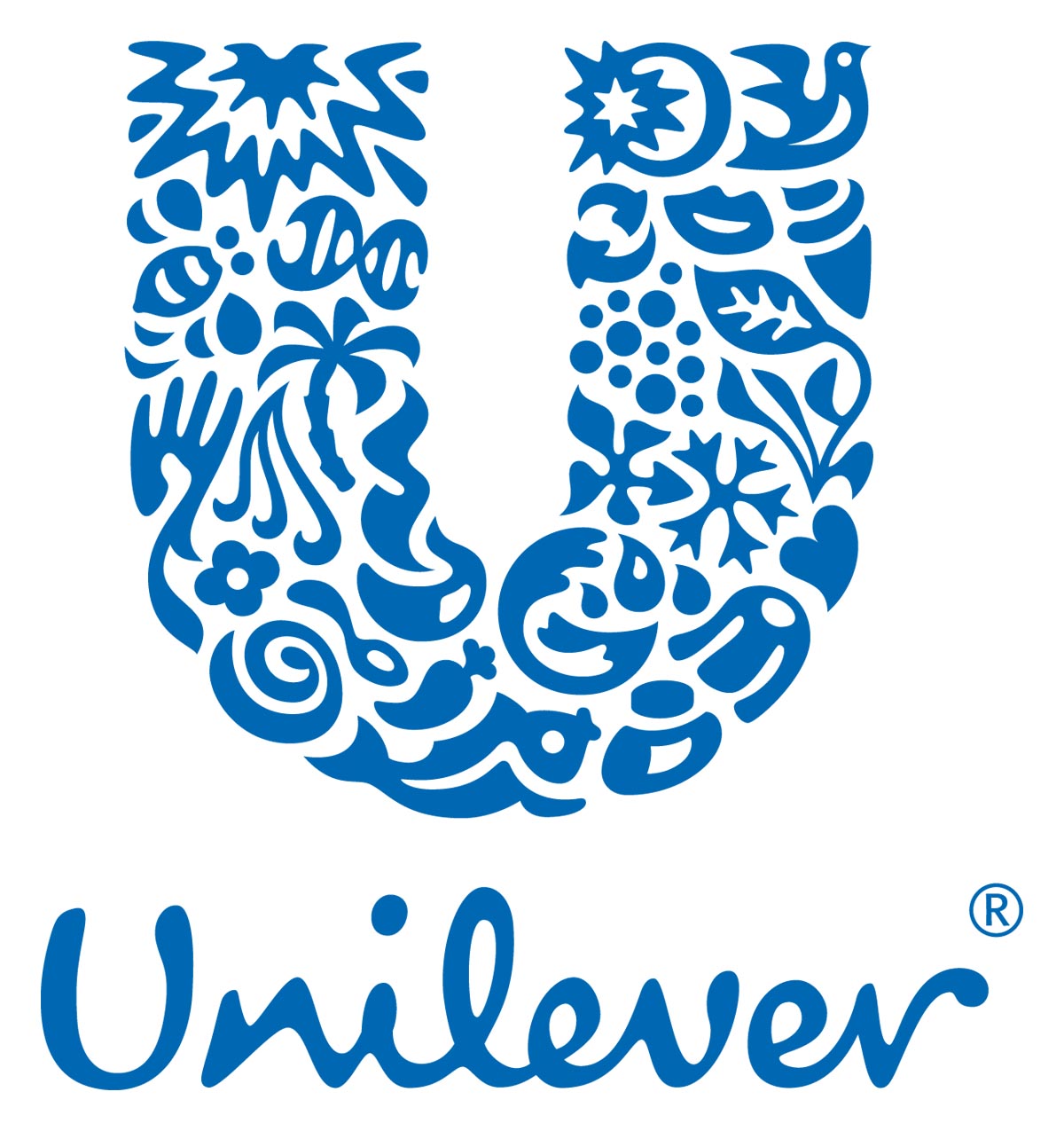}
		\includegraphics[width=0.065\textwidth]{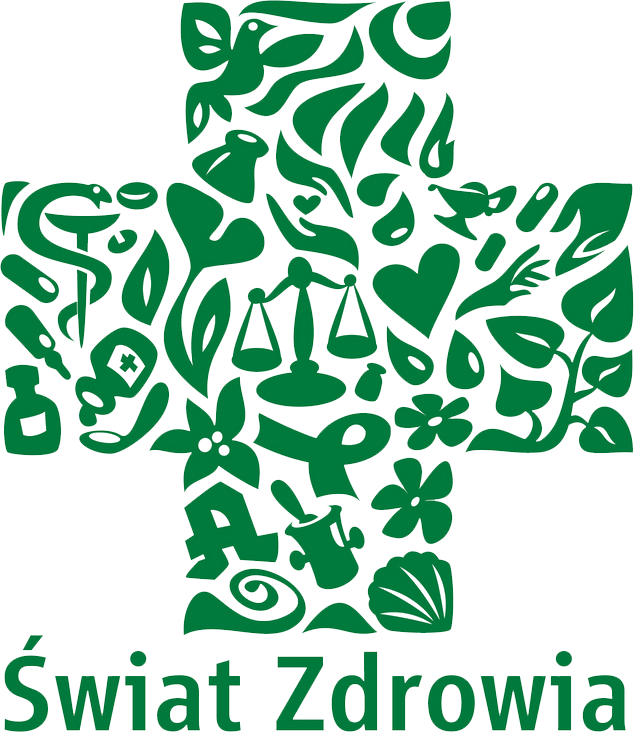}
	}
	\subfloat[Partial]{
		\includegraphics[width=0.07\textwidth]{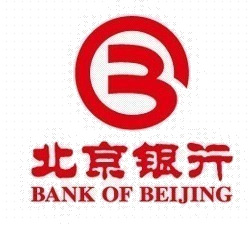}
		\includegraphics[width=0.07\textwidth]{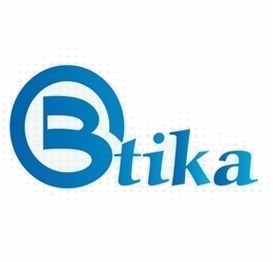}
	}
	\caption{Different types of similarity matches for trademarks.}
	\label{fig:vis_sam}
\end{figure}

Large-scale trademark retrieval (LSTR) studies return all trademarks that are similar or related to a given input from a set of at least one million trademarks. It is a challenging content-based image retrieval (CBIR) problem and as a result, it involves all of the challenges inherent in CBIR problems such as a large search space, partial/semantic similarity, and limited computing resources. It also faces some other difficulties unique to trademarks. Trademarks contain less information than natural images as they are often heavily stylized, and they do not contain the rich texture which is regularly found in natural image content. Additionally, they share common design elements such as characters and icons. Furthermore, the definition of trademark similarity is ambiguous and broad. It includes various aspects, namely shape, phonetic, semantic, layout, texture and partial aspects as shown in Figure \ref{fig:vis_sam}. Last but not least, industrial-level trademark datasets like the METU dataset \cite{metudeeptursun} include three types of trademarks: \textit{text-only}; \textit{figure-only}; and \textit{figure and text} trademarks. Considering this database, six different combinations of trademark types will appear during similarity calculation. It is challenging to apply a unified approach to precisely calculate similarities between different types of pairs. 
This further complicates an already extremely difficult large-scale retrieval challenge.



Recent studies \cite{metudeeptursun,lan2017similar,aker2017analyzing,feng2018aggregation} have utilized key-point features \ie SIFT \cite{lowe2004distinctive} and deep off-the-shelf DCNN features \cite{simonyan2014very} for LSTR. Although the results of these systems, especially those that used DCNN features, show significant improvement in retrieval performance compared to previous results, even the best retrieval rank is still below an acceptable level of accuracy (for example, \cite{metudeeptursun} achieves 0.063 normalized average rank). Investigation of the failure cases of these system show that the poor performance is a result of similarities caused by text, background, tiny symbols, contrast or noise.

\begin{figure}[!th]
	\centering
	\subfloat[VGG-16/FC7\cite{simonyan2014very}]{
		\includegraphics[width=0.5\textwidth]{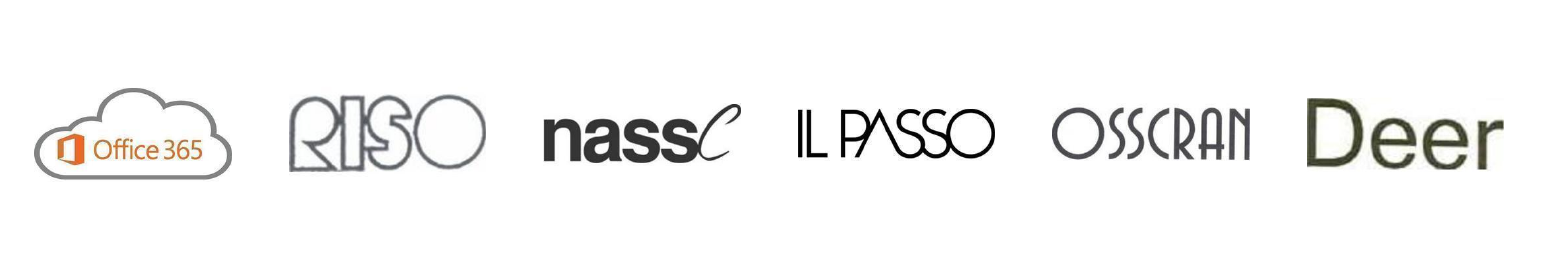}
	}
	\\
	\subfloat[Inception-Resnet-v2/PreLogitsFlatten\cite{szegedy2017inception}]{
		\includegraphics[width=0.5\textwidth]{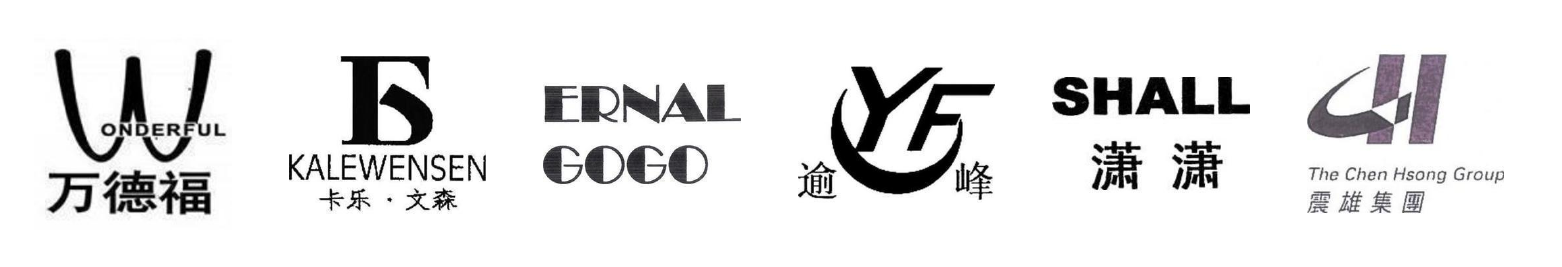}
	}
	\caption{Failure cases of trademark retrieval using off-the-shelf DCNNs. In each row, the image on the left is the query, and others are top five results from the METU trademark dataset.}
	\label{fig:problem}
\end{figure}

By far the most common failure cases, as shown in Figure \ref{fig:problem}, are related to text-elements in trademarks. 
This happens for the following reasons: firstly, similarities may exist between text elements in a pair of trademarks, but they are not considered infringements, as in general figurative similarities are more decisive than textual similarities since fonts, characters and symbols belong to public design elements; secondly, similarities also exist between figurative and text elements in a pair of trademarks, since textual parts have sharp corners, holes and strokes which are cable of generating rich features identical to some figurative elements. 

We therefore surmise that reducing the contributions of features from textual elements in trademarks when calculating similarity values should boost both existing deep and hand-crafted feature-based methods.



Our intuition is shown to be correct when considering the improvements in mean average precision (MAP) scores achieved on the METU trademark dataset \cite{metudeeptursun} via two naive approaches which are designed to increase attention given to figurative parts. In one experiment, through considering the dissimilarity between figurative and text elements, the similarities of pairs formed by a text-only trademark and a figure-only trademark are intentionally decreased (``rerank'' in Table \ref{tab:naive}). We decrease the similarity by an amount depending on the sum of the text elements appearing in them. In another experiment, we manually removed text from queries in the METU dataset (``crop'' in Table \ref{tab:naive}). The results of these naive methods, shown in Table \ref{tab:naive}, illustrate that directing attention away from non-informative or distracting parts of the trademark is a viable strategy. However, these naive methods have obvious limitations. Manual text removal lacks scalability, while the rerank method only has slight improvements even with very accurate labels.

\begin{table}[!htb]
	\begin{center}
		\begin{tabular}{c c c c}
			\hline
			\bf Model & \bf baseline &\bf rerank &\bf crop\\ 
			\hline\hline
			VGG-16/FC6 \cite{simonyan2014very}   	& 19.3   &  19.7  	& 27.3\\
			VGG-16/FC7 \cite{simonyan2014very}   	& 18.0   &  18.3  	& 24.7\\
			Inc.\_Res.\_v2 \cite{szegedy2017inception}  & 17.7   &  18.0  	& 23.7\\
			SIFT BoW \cite{metudeeptursun}	 & 16.5	 &  17.3    & -   \\\hline
		\end{tabular}
	\end{center}
	\caption{Examples of the effectiveness of directing attention toward figurative parts of trademarks when calculating similarities. The MAP@100 value is provided in this table.}
	\label{tab:naive}
\end{table}

In this paper we introduce hard and soft attention to automatically direct attention to critical information such as figurative elements for precise similarity calculation with off-the-shelf DCNN and hand-crafted features. For hard attention we have developed a novel text removal method to segment figurative portions of trademarks. It is a hybrid version of two popular text removal strategies: text detection plus inpainting\cite{esedoglu2002digital} and image translation\cite{nakamura2017scene}. For soft attention we investigate both fully and weakly supervised saliency maps. The fully supervised saliency map is generated by the same network that is used for the hard-attention, while the weakly supervised saliency map is generated via convolutional activation maps (CAM) \cite{zhou2016learning}.

Our main contribution in this paper is the development of techniques to direct attention toward critical information in the trademark to improve TR performance with the off-the-shelf CNN features. By proposing both hard and soft attention approaches, we have improved on the state-of-the-art performance on the largest and most challenging public trademark dataset: the METU trademark dataset\cite{metudeeptursun}. What's more, the feature dimensionality of our proposed system, which achieves new state-of-the-art results, is only 256, making the representation more efficient than other recent approaches (\ie \cite{metudeeptursun, aker2017analyzing, lan2017similar, feng2018aggregation}).

\section{Literature Review}
\label{sec:lit}
%


Recent state-of-the-art results for instance-level image retrieval benefit from aggregation and embedding \cite{babenko2015aggregating, ng2015exploiting, hoang2017selective, tolias2015particular, wei2017selective, kalantidis2016cross, gong2014multi}, PCA  whitening \cite{babenko2015aggregating, babenko2014neural, razavian2016visual, gong2014multi}, mid-level convolutional features \cite{babenko2015aggregating,ng2015exploiting,razavian2016visual}, fusion of convolution features \cite{zheng2016good}, multi-resolution searching \cite{razavian2016visual, gong2014multi}, fine-tuning \cite{babenko2014neural} and attention \cite{babenko2015aggregating, kalantidis2016cross, hoang2017selective, wei2017selective} approaches.

There exist two kinds of attention approaches: \textit{weighting} \cite{babenko2015aggregating, kalantidis2016cross} and \textit{selection} \cite{hoang2017selective, wei2017selective}. Weighting approaches create attention by emphasizing convolutional activations of relevant information or by reducing activation of irrelevant information via multiplying weights. SPoC \cite{babenko2015aggregating} assigns attention to the center region of the input by multiplying weights generated by a geographical center prior heuristic. CRoW \cite{kalantidis2016cross} proposed a non-parametric cross-dimensional weighting framework, which includes spatial and channel-wise weightings. The spatial weight values are based on the normalized total responses across all channels in a convolutional layer, while the channel whitening is based on the sparsity of the feature maps. Jimenez \etal \cite{Jimenez_2017_BMVC} aggregate attentive convolutional features of the top predicted classes to go beyond spatial attention. Their approach is similar to our weakly supervised soft attention approach in that both utilize convolutional activation maps (CAM) \cite{zhou2016learning} to boost the performance of off-the-shelf features for image retrieval; however our approach differs as follows: (1) We use a two-stream network, composed of a pre-trained network and a CAM network, instead of a single stream CAM-modified pre-trained network. It is found that the CAM-modified pre-trained models perform worse than the originals on both  classification and retrieval tasks \cite{zhou2016learning, Jimenez_2017_BMVC}. What is more, in our  experiments, we also found that fine-tuned pre-trained networks for trademark type classification perform worse than the original networks on the TR task; and (2) In our case, the prediction layers are removed from both pre-trained and CAM networks since most of the trademarks only include a single object, therefore, aggregation based on the top $N$ predicted classes is unnecessary.




%
%

Selection approaches direct attention to import information by selecting convolutional features; and the process is equivalent to applying a binary weight spatially in the case of using  global average pooling or maximum activation pooling as aggregation techniques. For example, Wei \etal \cite{wei2017selective} selected local features on the largest activated connected component of a convolutional layer. Hoang \etal \cite{hoang2017selective} select deep convolutional local features via masks (\ie Max, Sum, SIFT) and aggregate them to a global feature. 

The regional maximum activations of convolutions (R-MAC) \cite{tolias2015particular} shows state-of-the-art results with off-the-shelf features on several image retrieval baselines. R-MAC aggregates processed MACs of all regions generated by sliding different scale windows over the input image. Although implicit spatial weighting is applied by R-MAC when overlap exists between windows, taking each sliding window's importance into account is more promising, since the most discriminative regions belong to certain sliding windows. Kim \etal \cite{kim2018retrieval} improved on R-MAC by assigning learned weights to all regions.
%
%

Boosting retrieval performance of intermediate and late convolutional layers by multiplying saliency or semantic parsing is also useful for image retrieval problems. Recent studies \cite{kalayeh2018human, quispe2018improved} on person re-identification apply saliency and semantic parsing. Image geo-localization is another image retrieval problem that benefits from attention being placed on critical information. Kim \etal \cite{kim2017learned} learned visual attention with a contextual reweighting network. Another highlight of their work is that they applied attention scores to the aggregated convolutional features with VLAD aggregation \cite{arandjelovic2016netvlad}.

Although those attention mechanisms bring improvements to other image retrieval tasks, most are designed for natural images. Thus, performance in TR may be limited due to differences between natural images and trademarks. What's more, the notion of attention emphasizes task-relevant optimization. Compared to those existing approaches, our hard and soft attention is more specific to TR.

\subsection{Trademark Retrieval}
Tursun \etal \cite{metudeeptursun} compared various hand-crafted and DCNN features on the METU trademark dataset. Their results show that deep features are superior to hand-crafted features on the METU dataset, but fusion is necessary to achieve state-of-the-art results. They also noticed the low-accuracy caused by text elements, but their discovery and subsequent mitigation techniques are limited to local hand-crafted features. Aker \etal\cite{aker2017analyzing} provided further analysis on TR with deep features. Although their work shows that deep features are superior to hand-crafted features, deep features are also negatively affected by transformation, contrast, scale, and aspect ratio. Lan \etal \cite{lan2017similar} also utilized mid-level convolutional features extracted from a pre-trained network, however they applied uniform local binary patterns (LBP) to features maps for aggregation. Although their approach improved performance of off-the-shelf features, the aggregation process is costly and the aggregated features lack scalability because of large dimensionality. Feng \etal \cite{feng2018aggregation} extracted reversal invariant SIFT features from edges of the segmented blocks of a trademark, then aggregated SIFT features from each block to generate a single global representation. This method also lacks scalability, as multiple features are used for each trademark. 


\section{Proposed Approach}
\label{sec:met}
As discussed in Section \ref{sec:intr}, when the query includes both figurative and text components, more precise trademark similarity scores can be obtained by directing attention to informative regions, \ie figurative elements. To achieve this, we introduce hard (Section \ref{sec:hardatt}) and soft (Section \ref{sec:softatt}) attention approaches. They are able to direct extra attention to figurative regions as per the naive approaches, while being scalable, robust and efficient.

\subsection{Automated Text Removal Hard Attention (ATRHA)}
\label{sec:hardatt}
As noted earlier, removing text from figurative and text-figurative trademarks will improve retrieval results since features from figurative regions yield more robust features for retrieval. Approaches that direct focus towards certain regions by ignoring all others are known as hard-attention. We propose the ATRHA approach for TR. It automatically removes text on figure-only and figure and text trademarks before applying the feature extraction process. Therefore, the text removal step is key to the ATRHA approach.

There are two types of text removal approaches \cite{nakamura2017scene}: \emph{text detection plus inpainting} \cite{esedoglu2002digital} and \emph{image transformation} \cite{nakamura2017scene}. However, they are not suitable for reducing the amount of non-critical information for accurate similarity retrieval. The disadvantage of text-inpainting is the imprecise text localization \cite{tian2016detecting}. Object detection based text localization applies a bounding box for localization. Bounding box based methods require additional processes for segmenting text pixels, and highly stylized texts make this challenging. On the other hand, image translation techniques have low translation accuracies at the edge of text regions.

\label{sec:unet}
\begin{figure*}[!tb]
	\centering
	\includegraphics[width=0.9\textwidth]{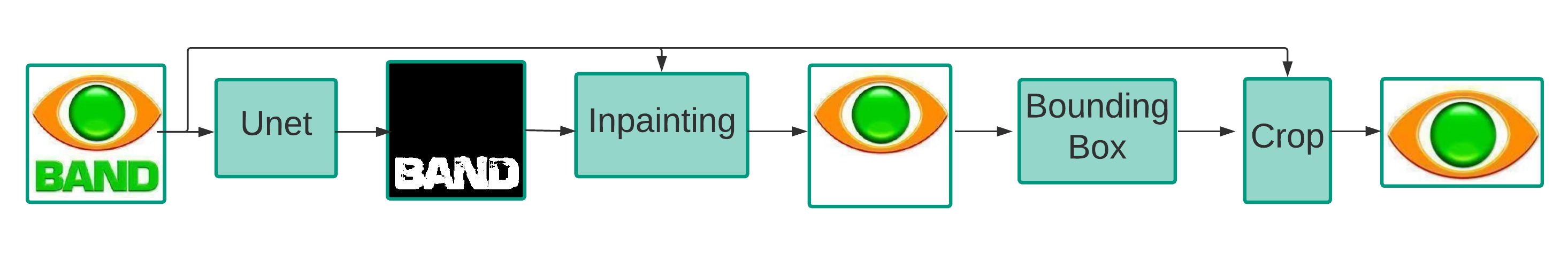}
	\caption{The proposed scheme for automatic text removal. U-Net segments the input into text and non-text pixels. Text pixels are removed by in-painting with their nearest dominant background colors. The bounding box of the foreground is extracted from the inpainted input. The input is cropped for text-removed input.}
	\label{fig:txtrm-scheme}
\end{figure*}
Our proposed text removal method is a combination of text-inpainting and image translation approaches with innovative additions. As shown in Figure \ref{fig:txtrm-scheme}, an image is segmented into text and non-text pixels. Then text pixels are inpainted based on the corresponding background pixels. Later, an auto-trim algorithm is applied to locate the bounding box of the non-text area. Finally, the trademark with removed text is obtained by cropping the given image with the obtained bounding box. Note the original input is returned when the bounding box is extremely small to avoid text removal on text-only trademarks.
The proposed approach uses the following components.

\textbf{U-Net:} U-Net\cite{ronneberger2015u} is developed for precise image segmentation tasks with few training images. Here, U-Net is used to separate pixels into text and not-text pixels. Its structure is similar to fully convolutional segmentation networks, which are composed of deep encoder and decoder networks. However, its decoder not only includes feature channels generated by up-sampling layers but also features from the encoder. Those shared feature channels allow the network to propagate context information, which increases the accuracy of the network and improves the model.
	
\textbf{Focal loss:} U-net is trained with the focal loss\cite{lin2017focal}, which is expressed as,
\begin{equation}
FL(p_t)=-(1-p_t)^{\gamma}\log(p_t),
\label{eq:fl}
\end{equation}
where $p_t$ is the probability of the input object belonging to class $t$, and $\gamma$ is a tunable focusing parameter. This loss emphasizes sparse text-pixels by adding a factor to the standard cross entropy criterion. In trademarks, the number of text pixels is often far fewer than the number of background and non-text pixels.
Therefore training datasets for pixel-wise text segmentation are highly imbalanced. Training with the standard cross entropy loss on this dataset will result in the model optimizing for the dominant class rather than all classes.

\textbf{Inpainting:} We develop a fast and efficient inpainting technique, exploiting the nature of trademark design. Usually, trademarks have monotone backgrounds. Therefore the most dominant color around non-text pixels is identified as the background color. Inpainting is achieved by replacing text-pixels with the corresponding background pixel. When identifying background color, a morphological closing operation is applied to the mask image from U-Net to correct false negatives around text-pixels. 

\textbf{Cropping:} Bounding box detection and cropping processes follow the inpainting process to fix potential errors left from previous steps. Their failures will bring unexpected negative changes to the appearance of a given trademark. To keep trademark appearance as original as possible whilst erasing text, the original trademark is cropped with the bounding box of the inpainted trademark. However, a side-effect of the cropping is that text surrounded by figurative elements will remain.

\begin{figure}[tbh]
	\centering
	\includegraphics[width=0.45\textwidth]{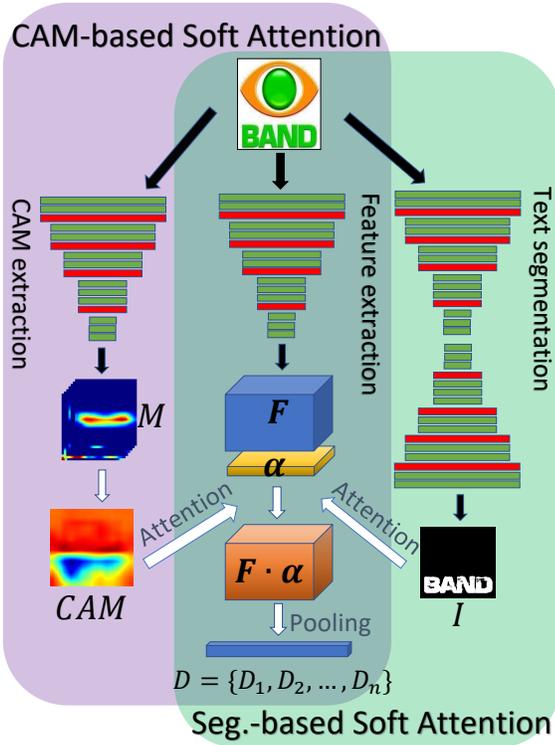}
	\caption{The pipeline of soft attention approaches for TR. Both proposed CAM-based (pink region) and Segmentation-based (green region) approaches are given in this figure. In both approaches, feature and saliency extraction is completed by separate networks.}
	\label{fig:soft-attention}
\end{figure}

\subsection{Soft Attention}
\label{sec:softatt}
%
A different approach to identify important regions is soft attention. It amplifies or decreases the role of features in similarity calculations by assigning weights to them based on heuristic or learned weight functions. Usually, soft attention is applied to convolution features, since convolution features keep spatial information. More formally, the feature $f_k(x,y)$ presents the activation of unit $k$ in a convolutional layer at spatial location $(x,y)$. The attention magnitude at the same location is the attention score $\alpha(x,y)$. Therefore, the attentive convolutional activation $f'_k(x,y)$ is,
\begin{equation}
f_k'(x,y) = \alpha(x,y)f_k(x,y).
\label{eq:atnpool}
\end{equation}

Compared to a hard-attention approach, soft-attention has the following advantages: (1) The attention mechanism can be integrated into the feature extraction pipeline; (2) Text information is preserved, albeit with a reduced weight, while figurative information is still the main focus during similarity calculation. This means that text that is within figurative regions can be more easily suppressed than is possible with hard attention.

\begin{figure}[!tbh]
	\centering
	\subfloat[]{
		\includegraphics[width=0.1\textwidth]{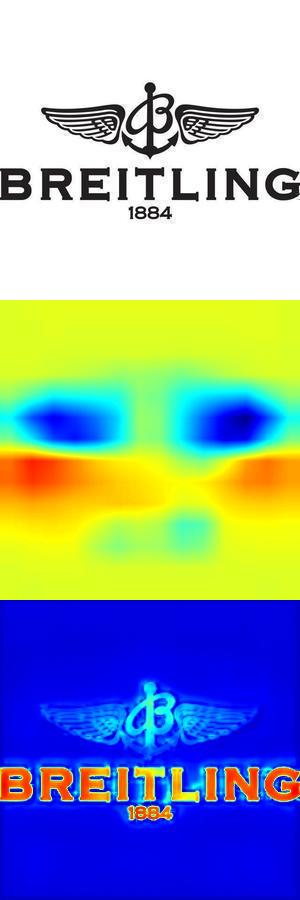}
	}
	\subfloat[]{
		\includegraphics[width=0.1\textwidth]{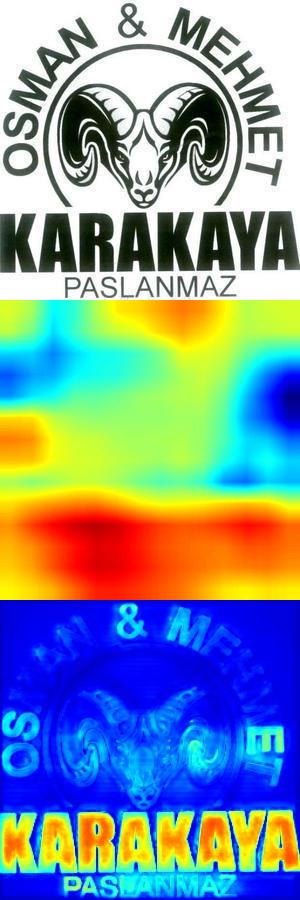}
	}
	\subfloat[]{
		\includegraphics[width=0.1\textwidth]{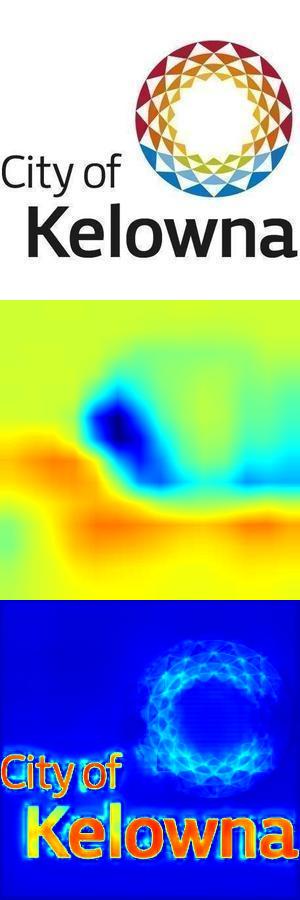}
	}
	\subfloat[]{
		\includegraphics[width=0.1\textwidth]{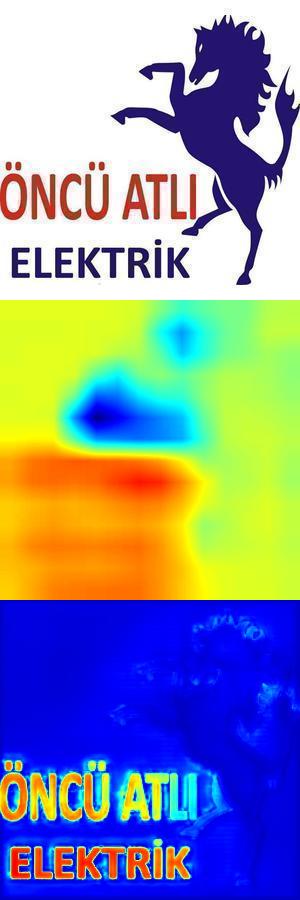}
	}
	
	\caption{Examples of text-related saliency maps generated by CAM and U-Net. Each example presents the image, its CAM and U-Net based saliency maps from top to bottom. Saliency maps are pseudo-colored from red to blue depending on the intensity. In CAM, red corresponds to text (low attention), while blue corresponds to figurative (high attention). In U-Net, red and yellow are text, while blue is the background.}
	\label{fig:cam-unet}
\end{figure}

We propose two approaches for generating soft attention scores for TR. One approach is based-on the U-Net architecture used for hard attention. Another approach is based on the CAM. As illustrated in Figure \ref{fig:soft-attention}, the pipelines of two approaches are similar, though they generate distinct saliency maps. As shown in Figure \ref{fig:cam-unet}, the CAM generates attention weights for both figurative and text parts. In comparison, U-Net returns precise attention for text but fails to discriminate figurative regions from the background.

\subsubsection{Segmentation-based Soft Attention (SSA)}
\label{sec:seg}
As hard attention is a simple case of soft attention, the U-Net segmentation mask is converted to soft attention weights. If a weight of 0 is assigned to the regions to be ignored (\ie text), this soft attention scheme becomes an approximation of a hard attention scheme. Inverting this, the segmentation mask, $I$, used by hard-attention can be converted to a soft attention score $\alpha$ as follows,

\begin{equation}
\alpha(x,y) = 1 + b - I(x,y).
\end{equation}

In experiments, $b$ is set to 0.5; $\alpha$ for text pixels lies in range 0.5 to 1, and 1 to 1.5 for non-text pixels.

\subsubsection{CAM-based Soft Attention (CAMSA)}
\label{sec:cam}
We also propose a CAM based attention score. Comparing to the U-Net segmentation approach, CAM is more efficient in terms of training requirements.

The work by Zhou \etal \cite{zhou2016learning} shows that CNNs trained for classification can be used for object localization when the fully connected layers are replaced by a global average pooling layer. Their method is known as CAM (convolutional activation map). The global average layer enhances feature maps in the last convolutional layer to learn object locations when training for classification. Therefore, the weighted sum of those feature maps corresponds to an object. More precisely, the predicted class will have high attention when the weights correspond to the predicted class.

Suppose that we trained a network for classifying a trademark into three classes: text-only, figure-only, and figure and text. The CAM for any class can be mapped to an attention score as follows,

\begin{equation}
M_i(x,y) = \sum_{k} w_k^if_k(x,y),
\end{equation}

where $i$ is the class index, and $w^i_k$ is the weight corresponding to class $i$ for unit $k$.

In experiments, the CAM for the figure-only class shows the best results. The function that maps the CAM for the figure-only class to an attention score is defined as follows,

\begin{equation}
	\alpha(x,y) = \beta^tM_i(x,y),
\end{equation}

where $t$ is 1 when $M_f(x,y) > \tau$, and otherwise -1. In experiments, $\beta$ and $\tau$ are set to 100, 0.5 respectively.


\subsubsection{Feature Aggregation}
Aggregation of convolution features is necessary for compact representations. In general, to generate a global feature $F$, aggregations such as global average pooling/sum pooling (SPoC), or maximum activation pooling (MAC) are applied to each unit, $k$, of a convolutional layer. Formally, SPoC and MAC pooling for attentive features are expressed as,

\begin{equation}
s'_k = \sum_{y}^{H}\sum_{x}^{W}f'_k(x,y),
\label{eq:sum}
\end{equation}
\begin{equation}
m'_k = \max_{x,y}f'_k(x,y).
\label{eq:max}
\end{equation}

In general, an $l_2$ normalization is applied to a global feature $F=\{f'_1, f'_2, \dots, f'_n\}$. For SPoC, following $l_2$ normalization, PCA-whitening with dimension reduction and $l_2$ normalization are applied to further reduce the memory, feature burstiness and improve robustness to noise.

\section{Experiments}
\label{sec:exp}

\subsection{Datasets}
\textbf{METU Trademark Dataset.} The METU dataset \cite{metudeeptursun} is the largest public dataset for TR \cite{tursun2015metu}. It includes $589,098$ text-only marks, $19,387$ figure-only marks and $311,986$ figure and text marks. Its evaluation set is composed of 417 queries with the expected similar marks. 
%
In experiments, three modifications are introduced to the METU trademark dataset. The first is that trademarks are resized to 300$\times$300, but their aspect ratio is maintained by padding with white space. Secondly, a new evaluation set is created by manually removing text regions of the default validation set. 
This evaluation set is used for the manual text removal experiment of Section \ref{sec:intr}. Lastly, a new dataset is generated by automatically removing text regions on all trademarks using the text removal method proposed in Section \ref{sec:hardatt}.

\textbf{Pixel-level Text Localization (PTL) Dataset.} The PTL dataset is created for training U-Net as discussed in Section \ref{sec:unet}. It is mainly composed of 100,000 synthetic images, generated by randomly placing text on backgrounds with or without figurative elements. Note that all text, backgrounds and figurative elements are randomly generated or selected. Additionally, the PTL dataset includes 12,000 text-only trademarks with a white\footnote{RGB(255,255,255)} background and 9,000 figure-only trademarks from the METU trademark dataset, making the dataset more realistic and balanced. 
The ground truth mask images are obtained by comparing normalized images with their backgrounds at the pixel-level. If the difference is less than $0.05$, the pixel is not part of the text.

\textbf{Trademark Type (TT) Dataset.} The TT dataset is prepared for fine-tuning CNNs, which classify trademarks for generating CAM as explained in Section \ref{sec:cam}. It consists of 12,000 METU trademarks for each trademark type, while its validation set includes 2,000 trademarks in total.

%

\subsection{Evaluation}
\label{sec:eval}
The performance of the proposed algorithms are evaluated using \textit{mean average precision} (MAP) of the top 100 results. The ranking is achieved by sorting the similarity values of a query and other trademarks in descending order. In the event that two trademarks, one that is similar and one that is not, match the query with equal accuracy, we order them such that the similar query is ranked behind the other (\ie we take a pesamistic view).


The similarity of a pair of trademarks is equal to one minus the distance between their $l_2$ normalized features. Conventionally, the Euclidean distance is calculated for deep features, while the cosine distance is calculated for SIFT features.

\subsection{Network Parameters and Training}
In experiments, a pre-trained VGG-16 (trained on ImageNet) is used as the network for feature extraction. Features are extracted from $conv5\_3$.

In addition to feature extraction, VGG-16 is also modified and fine-tuned for extracting CAM. Its last fully connected layers are replaced with a global average pooling layer and a fully connected layer with 3 classification nodes. For clarity, we refer to this architecture as VGGCAM-16. VGGCAM-16 is fine-tuned on the TT dataset. The RMSprop optimizer is used with a learning rate 0.001, and batch-size of 32. The training process is stopped after one epoch, since the classification accuracy on the TT evaluation set has reached 96.65\% at this point.

For the hard-attention approach, U-Net is trained on the PTL dataset. U-Net is the same network proposed by \cite{ronneberger2015u} except for the input image size. Our input size is 256$\times$256. During training, the focal loss is applied. Its $\alpha$ and $\gamma$ values are set to 0.25, 2 respectively. The Adam optimizer is selected because of fast convergence. The initial learning rate is 0.001 and decays every 20,000 steps with a decay factor of 0.95. The batch size is 8. After 8 epochs, the F1 score on the PTL test set is 91.75\%. For further regularization, batch normalization (decay=0.997 and epsilon=0.001) and dropout (probability=0.9) are applied. Additionally, data augmentations such as rotating, flipping, shearing, stretching and translation are applied.

\newcommand{\tabfig}[1]{  
	\raisebox{-0.8\totalheight}{\includegraphics[width=0.055\textwidth]{images/qualitative_results/#1}}}

\begin{table*}[!t]
	\centering
	\begin{tabular}{c  l  c  c  c  c  c  c  c  c  c  c}
		\hline
		&  \bf Index &   1 & 2 & 3 & 4 & 5 & 6 & 7 & 8 & 9 & 10 \\ \hline\hline
		& \textbf{Query} & \tabfig{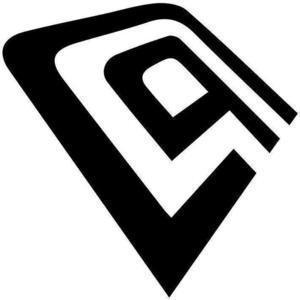}	& \tabfig{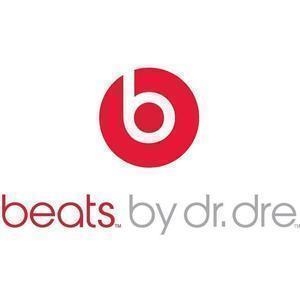} & \tabfig{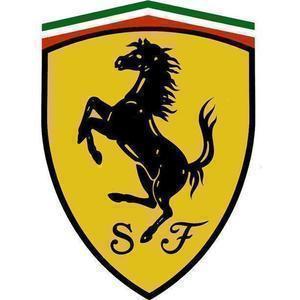} & \tabfig{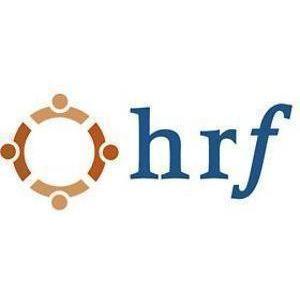}  &  \tabfig{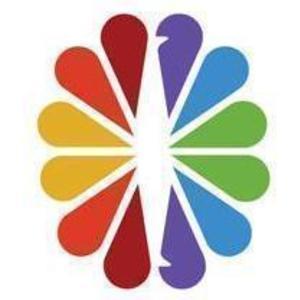} & \tabfig{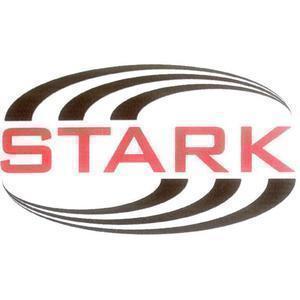} & \tabfig{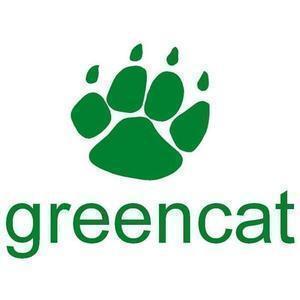} & \tabfig{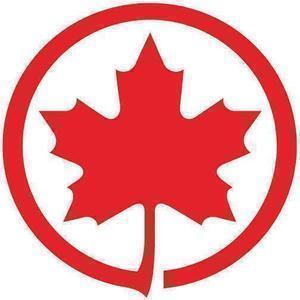}  & \tabfig{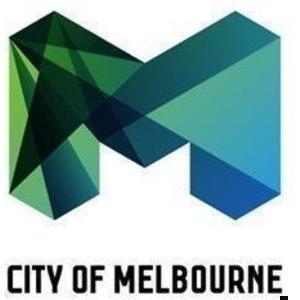} & \tabfig{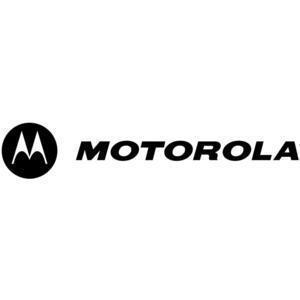}    \\
		& \textbf{Expected} & \tabfig{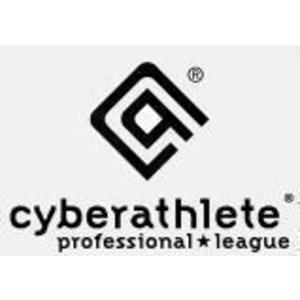}	& \tabfig{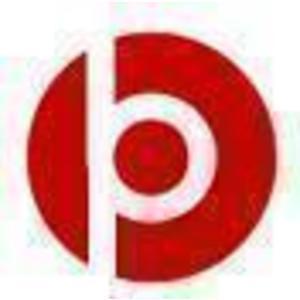}     & \tabfig{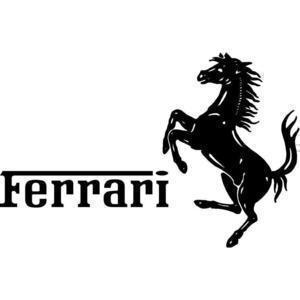}    & \tabfig{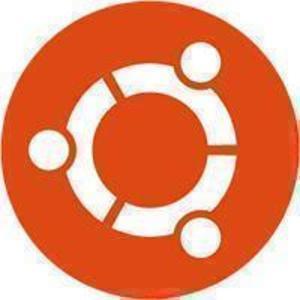}   	  & \tabfig{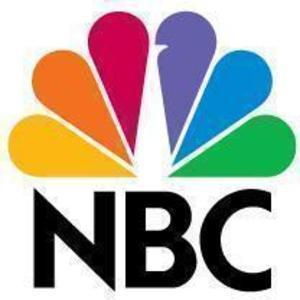}      & \tabfig{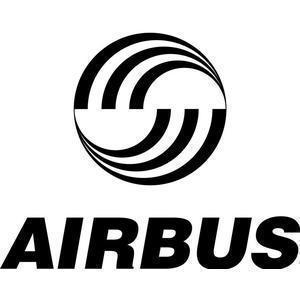} & \tabfig{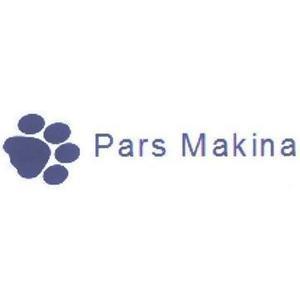} & \tabfig{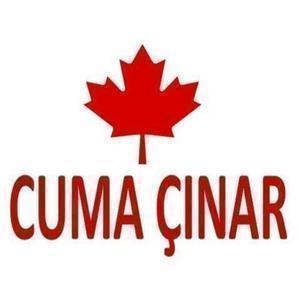} & \tabfig{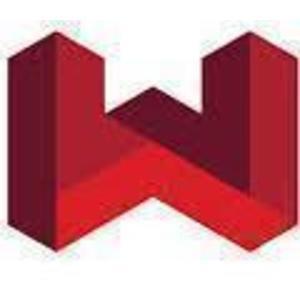} & \tabfig{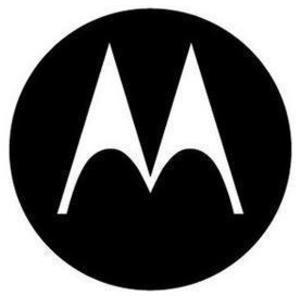}    \\ \hline\hline
		\multirow{6}{*}{\rotatebox[]{90}{\bf Baseline}} 
		& SIFT BoW \cite{metudeeptursun} &  78k   & 304k    &   78 	  &  779k     & 469 &   335k   & 165k  & 14k    & 316  &  4\\
		
		& MAC \cite{tolias2015particular}		&  31k    &  600k     &   81 	  &  324k     & 103 &   2k   & 39k  &  5k     & 302  &  893k\\ 
		& SPoC PCAw \cite{babenko2015aggregating} 		&  16k    &  28k     &   11 	  & 12k      &  15k &   206    &  32k  &   7k    &  1k & 253k \\ 
		& CRoW \cite{kalantidis2016cross} 		& 35k     &  49k     &   \bf 7	  &   1k    &  1k &  309     &  2k  &  \bf 646     &  42   & 381k  \\ 
		& Jimenez \cite{Jimenez_2017_BMVC} 		& 314     &   66k    &   419 	&  278   &  \bf 3  &  1k &  3k     & 34k &     52  & 42k \\ 
		& R-MAC \cite{tolias2015particular} 		&  1k    &  33k    &   10 	  & \bf 39      &  5 &  108     &   2k  &   2k    &  75 &   687k\\ \hline\hline
		\multirow{7}{*}{\rotatebox[]{90}{\bf Ours}} & ATRHA MAC 		&  \bf 5    &   \bf 3    &   1k &  2k     &  92 & 4k      &  99  &  15k     &   4  &  4 \\
		& ATRHA SPoC 		& 13     &  26     &  18  &  52k   &  937 &  588 &  43 &   3k    &  \bf 3   &  6 \\  
		& SSA MAC & 3k     & 533k      &  2k  &  54k     &  65 &   649    &  55k  &   31k    &  11 &  848k \\ 
		& SSA SPoC & 379     & 150k      & 12     &  134    & 281  &    \bf 38   &   8k & 2k     &  831  & 366k\\ 
		& CAMSA MAC  &  109    &  44k     &    199	  &    55   & 7  &   274    &  1k &   24k    & 65  &   727k\\ 
		& CAMSA SPoC		&   301   &    6   &  136  &   53    &  218 &  675     & 2k   & 873      & 219 & 404k\\ \cline{2-12}
		& ATRHA R-MAC		&   7   &    \bf 3   &  18  &   72k    & 17  &   303    & 41   & 456      & \bf 3 & 5\\
		& ATR. CAM. MAC		&   \bf 5   &   \bf 3   &  89  &   521    &  16 &  685     & \bf 7   & 15k      & 6 & \bf 3\\ \hline
		\hline
	\end{tabular}
	\caption{Rank-based qualitative results. Ranking results of 10 pairs of query and expected trademarks with various attention methods are compared. The best rank is one, and large ranks are rounded to the nearest thousand.}
	\label{tab:qual}
\end{table*}

\subsection{Feature Extraction}
Our feature extraction process is similar to general feature extraction with deep networks. However, some critical details are illustrated below.


\textbf{Center-crop.}
All features are extracted without applying center-cropping to the input image, as experimental results show that it is harmful to retrieval. For example, in Table \ref{tab:crop}, results of MAC and SPoC with and without center-cropping are presented. Comparing to SPoC, MAC is more sensitive to center-cropping, because cropping will remove certain maximum activations, yet can't substantially alter the average activation value.

\begin{table}[!h]
	\begin{center}
		\begin{tabular}{c c c}
			\hline
			\bf Method & \bf Crop & \bf MAP\\ 
			\hline\hline
			 \multirow{2}{*}{MAC\cite{tolias2015particular}} & - & 0.215 \\
												& \checkmark & 0.182 \\\hline\hline
			\multirow{2}{*}{SPoC\cite{babenko2015aggregating}} &  - & 0.177 \\
												&  \checkmark & 0.166  \\\hline
		\end{tabular}
	\end{center}
	\caption{Examples of effect of center-cropping on TR}
	\label{tab:crop}
\end{table}

\textbf{Feature Post-processing.}
Results of related works \cite{babenko2015aggregating,kalantidis2016cross, Jimenez_2017_BMVC,tolias2015particular} show the features post-processed with $l_2$-normalization, PCA-whitening and $l_2$-normalization have better performance especially when SPoC aggregation is applied. We reach similar findings for SPoC. However, MAC aggregation is negatively affected by this process. We refer to this process as ``PCAw''. To calculate PCA, we randomly selected 20,000 samples and retained the most important 256 components.

\subsection{Comparison to State-of-the-Art Results}

The proposed attention approaches are compared with both previous state-of-the-art results for TR and image retrieval. Results are given in 
\cref{tab:qual,tab:crop,tab:nar,tab:quan}. In these Tables, for convenience, we refer our proposed approaches with a concatenation of names that denote the applied attention, aggregation and processing methods. For example, the ATRHA with SPoC aggregation and PCAw process is referred as ``ATRHA  SPoC PCAw''.

Previous TR results are evaluated with the normalized average rank (NAR) \cite{metudeeptursun,feng2018aggregation} metric, and are shown in Table \ref{tab:nar}. The ATRHA CAMSA MAC method achieves new state-of-the-art results, and the ATRHA R-MAC approach shows results comparable to the previous state-of-the art. However, the previous state-of-the-art result is a fusion of both hand-crafted and DCNN features, and the minimum dimension size is 4,096. In comparison, both ATRHA CAMSA MAC and ATRHA R-MAC have compact feature vector dimensions of 512 and 256 respectively. Furthermore, a similar fusion process will enlarge the margin between our methods and previous state-of-the-art results as from Table \ref{tab:qual} it can be seen that our proposed methods have different precisions for different queries, suggesting complementary information that could be fused.

Our attention approaches are also compared with state-of-the-art results for image retrieval with off-the-shelf DCNN features: R-MAC, SPoC, CRoW and the method proposed by Jimenez \etal \cite{Jimenez_2017_BMVC}, in Table \ref{tab:qual} and \ref{tab:quan}. Their implementations are as per the original proposal. However, in R-MAC the scale parameter is set to 4; and for \cite{Jimenez_2017_BMVC} the top 8 predicted classes are considered for feature aggregation.

Results given in Table \ref{tab:quan}, show conclusively that attentive DCNN features increase performance except for SPoC PCAwc, which applies the central prior attention mentioned in \cite{babenko2015aggregating}. This clearly demonstrates that proper learned attention mechanisms are useful, especially when tailored to their application (TR in our case). ATRHA R-MAC returned the best MAP@100 score. 
 
Additionally, the results of our soft attention methods outperform other soft attention approaches except for R-MAC. In particular, CAMSA shows better results than SSA. Through comparing results of MAC and R-MAC in Table \ref{tab:qual}, R-MAC has a huge improvement over MAC, especially scale changes are present. However, R-MAC is not accurate when text included in trademarks. To overcome it, R-MAC is used with ATRHA, and it result surpassed other methods by a large margin. In comparison, our soft attention approaches lack scale-invariance. Although it is also true for our hard attention, the hard attention overcomes scale problems with text removal in some cases. For example, the 10th case in Table \ref{tab:qual} contains a large scale change. Most of them fail because of the scale difference. However, ATRHA based methods bypass the scale problem by removing text and resizing, and SIFT is scale invariant. Therefore, our soft attention approaches should incorporate scale invariance for further improvement. This idea is supported by the results of ATRHA CAMSA MAC, which is a combination of our soft and hard attention methods.

Apart from the attention, the selection of aggregation methods is also important. MAC is superior to SPoC. However, the PCAw process decreases the performance of MAC, therefore, its feature dimension is 512, twice that of SPoC.

The above comparison concerns quantitative results. Qualitative results for ten pairs of similar trademarks are given in Table \ref{tab:qual}. In most cases, attention methods improve results dramatically. However, these improvements are not completely captured by the MAP@100 score because of the limit on top results.  

\begin{table}[t]
	\begin{center}
		\begin{tabular}{c  l  c  c}
			\hline
			& \bf Method & \bf NAR & \bf MAP \\ \hline\hline
			\multirow{2}{*}{\bf Baseline} & Feng \etal \cite{feng2018aggregation} &  0.083 &  - \\
			& Tursun \etal \cite{metudeeptursun}   &  0.062 &  - \\\hline\hline
			\multirow{2}{*}{\bf Ours} & ATRHA CAMSA MAC 								  & \textbf{0.040} &  25.1 \\
			& ATRHA R-MAC 							 & 0.063	&  \textbf{25.7} \\\hline
		\end{tabular}
	\end{center}
	\caption{Comparison with the previous state-of-the-art results on METU dataset. NAR is the normalized average rank metric. A smaller NAR indicates better results.}
	\label{tab:nar}
\end{table}
\begin{table}[t]
	\begin{center}
		\begin{tabular}{c l c c}
			\hline
			& \bf Method & \bf Dim. & \bf MAP\\ 
			\hline\hline
	\multirow{7}{*}{\rotatebox[]{90}{\bf Baseline}} & SIFT BoW \cite{metudeeptursun}	& 10k  &  16.5 \\
			& MAC\cite{tolias2015particular} 	& 512  & 21.5 \\
		    & SPoC PCAw \cite{babenko2015aggregating} 	& 256 	& 18.9 \\
			& SPoC PCAwc \cite{babenko2015aggregating} 	& 256  	& 16.5  \\
			& CRoW \cite{kalantidis2016cross} 			& 256   & 19.8      \\
			& Jimenez \cite{Jimenez_2017_BMVC}  			&  256 	& 21.0     \\
			& R-MAC \cite{tolias2015particular}  		& 256 	& 24.8 \\\hline\hline
	\multirow{8}{*}{\rotatebox[]{90}{\bf Ours}}		& ATRHA MAC								& 512   & 24.9 \\				
			& ATRHA SPoC    & 256    &  22.5 \\
			& SSA MAC   & 512  	& 21.3	\\
			& SSA SPoC PCAw & 256  	& 20.3	\\	
			& CAMSA MAC  & 512  & 22.3\\
			& CAMSA SPoC PCAw & 256  & 21.5\\\cline{2-4}
			
			& ATRHA R-MAC & 256    & \textbf{25.7} \\
			& ATRHA CAMSA MAC & 512    & 25.1 \\\hline
			
		\end{tabular}
	\end{center}
	\caption{Comparison of previous state-of-the-art results and attention-based methods.}
	\label{tab:quan}
\end{table}
\section{Conclusion}
\label{sec:con}

We have presented component-based hard and soft attention solutions for boosting the performance of a DCNN based large-scale trademark retrieval system. Our hard attention based approaches for LSTR have achieved new state-of-the-art results, outperforming previous works by a significant margin (both on MAP@100 and NAR scores), on the largest and most challenging trademark dataset. Meanwhile, our soft attention based approaches have also outperformed other similar works, confirming our intuition that directing attention to critical information in the trademark is a promising approach that can significantly enhance the performance of TR systems. Moreover, the feature dimensionality of our new state-of-the-art method is only 256, which is a more efficient representation when compared to larger dimensions used in previous works. Additionally, qualitative results indicate that our proposed techniques are robust in that they exhibit invariance to scale, rotation and reversal, which is also a very important and desired characteristic for effective trademark retrieval. In our current work, we have demonstrated the effectiveness of our attention based techniques using off-the-shelf DCNN for features extraction. In future work, research into new end-to-end deep learning models and fine-tuning pre-trained deep models with soft attention mechanisms will be pursued, to further improve the performance of LSTR.

{\small
\bibliographystyle{ieee}
\bibliography{egbib}
}

\end{document}